\documentclass{article}

\usepackage[latin1]{inputenc}
\usepackage{amssymb,amsmath,amsthm,array}

\usepackage{hyperref}
\usepackage{booktabs}
\usepackage{multirow}
\usepackage{wrapfig}
\usepackage{xspace}
\usepackage{standalone}
\usepackage{tikz}
\usepackage{caption}

\usetikzlibrary{arrows.meta}

\newtheorem{theorem}{Theorem}[section]
\newtheorem{corollary}[theorem]{Corollary}
\newtheorem{lemma}[theorem]{Lemma}
\newtheorem{proposition}[theorem]{Proposition}
\newtheorem{remark}{Remark}

\newcommand{\R}{\mathbb{R}}

\newcommand{\E}{\mathbb{E}}

\newcommand{\cL}{\mathcal{L}}
\newcommand{\C}{\mathcal{C}}
\newcommand{\D}{\nabla}

\newcommand\inner[2]{\langle #1, #2 \rangle}

\newcommand{\method}{ROOTS\xspace}
\newcommand{\methodiso}{ROOTS-iso\xspace}
\newcommand{\F}{\tilde{F}}

\begin{document}
\title{ROOTS: A passive approach to \\ Growing Neural Networks}

\author{Ying Jin$^{1}$, Yena Chang$^{2}$, Th\'{e}o Rudkiewicz$^{1}$, \\ Guillaume Charpiat$^{1}$ and Alex Davey$^{1}$
\thanks{Submitted to ESANN 2027. This work was supported by the EU under GA no.101135782 (MANOLO~project).}
\vspace{.3cm}\\
\footnotesize ${}^1$Universit\'{e} Paris-Saclay, CNRS, INRIA, LISN, 91405, Orsay, France
\vspace{.1cm}\\
\footnotesize ${}^2$EPFL, Lausanne, Switzerland\\
}

\date{}

\maketitle

\renewenvironment{abstract}{%
    \small
    \begin{quote}
    \noindent\textbf{\abstractname.}\enspace\ignorespaces
}{%
    \end{quote}
}
\vspace{-0.6cm}
\begin{abstract}
  In deep learning, we almost always use an architecture's default channel widths. We revisit this assumption, treating the channel widths as the solution of a constrained optimisation problem, minimising the training loss subject to a budget on a complexity (such as FLOPs), and solve it using growing networks, which adaptively enlarge small seed architectures over the course of training. Previous growing network methods try to predict the value of new neurons at birth, using gradient or curvature estimates. Viewing growth as reversed pruning, we estimate each neuron's contribution to the loss using only the existing training steps. Our method \method can find a near-optimal Pareto front of architectures, including those which pareto-dominate a standard ResNet, in comparable or less time than a single standard training run.
\end{abstract}

\section{Introduction}\label{sec:intro}

The width of each layer is one of the first choices in the design of a network, but it is something we rarely change when using an off-the-shelf architecture. For example, ResNets double the number of channels whenever the resolution halves~\cite{he_deep_2016}. In practice, we are interested not just in architectures which maximise performance, but in those which also satisfy some constraints on the complexity, such as the number of parameters or FLOPs.
Given a class $\mathcal{A}$ of architectures and the corresponding weight space $\Theta_A$, we wish to solve the following constrained objective
\begin{equation}
    \operatorname*{argmin}_{A\in\mathcal{A}}\min_{\theta\in\Theta_A}\ \cL(f_{A,\theta})\quad\text{subject to}\quad \C(A)\le \C_\textrm{max},
  \label{eqn:constrained}
\end{equation}
where $\C$ measures the complexity (FLOPs in this work) of architecture $A$, and $\C_\textrm{max}$ is the budget. Solving~\eqref{eqn:constrained} by manual trial-and-error requires a full training run for every candidate width configuration. Neural Architecture Search (NAS)~\cite{zoph_neural_2017} automates this search at a high cost: typically training many candidates before the selected architecture is trained from scratch. Growing neural networks~\cite{davey_survey_2026} propose an alternative approach by adaptively enlarging the architecture throughout the course of training. In particular, a growth method must decide at every growth event where the next units of complexity are best spent.
\paragraph{Related work.}
We refer the reader to~\cite{davey_survey_2026} for a recent survey on growing neural networks. NORTH~\cite{maile_when_2022} grows layers once their activation rank has exceeded a threshold. Neuron-splitting methods such as Energy-Aware Splitting~\cite{wang_energy-aware_2020} split a single neuron's weights into two, choosing which to split by the expected second-order change in the loss, subject to a budget constraint such as FLOPs. Also closely related are SENN~\cite{mitchellSelfExpandingNeuralNetworks2023}, which predicts future loss decrease in the Fisher geometry, and NAGN~\cite{wang_where_2026} decides where to grow from a normalised gradient norm, related to the isotropic limit of our criterion (Remark~\ref{rem:gradmax}).

Predicting the long-term benefit of a newly-grown neuron is much harder than evaluating that of existing ones. A new neuron's utility lies in what training will make of it, and first- or second-order quantities along this new direction are a poor guide, especially as the new neuron's weights will change rapidly in the first steps of its training. Instead of trying to predict future loss improvements (as many growing methods do), our method \method (Retrospective Observation Of Training Signals) tracks the historical contribution of existing neurons, and uses this to determine where to grow. 
This requires finding quantities which are meaningful to integrate over time.
Our approach is passive: it estimates curvature from the quantities that ordinary training already computes, with no additional forward or backward passes.

\section{Assigning credit to a neuron}

Let $f: X \to \R^K$ be the network's realised function, and for any example $x \in X$, the loss is $\ell_x: \R^K \to \R$, with expected loss $\cL = \E_x[\ell_x]$. Consider a quadratic approximation to the loss $\mathcal L$, which for parameters $\theta\in\R^d$ and $\delta \in \mathbb R^d$ gives
\begin{align}
  \cL(\theta + \delta) \approx \cL(\theta) + D_\theta(\delta), \qquad D_\theta(\delta) := \inner{\D_\theta\cL}{\delta}+\tfrac12\|\delta\|_{\F}^2
  \label{eqn:quadratic}
\end{align}
where $\|\delta\|_{\F}^2 = \delta^\top \F \delta$ is a semi-norm and $\F = \E_x[\D_\theta \ell_x \D_\theta \ell_x^\top]$ is the empirical Fisher matrix. It is an approximation of the true Fisher matrix \cite{martens_insights_2020}, but it can be computed from the ordinary training gradients without additional forward or backward passes (Section~\ref{sec:estimators}).

Under ordinary gradient descent, the parameters $\theta \in \R^d$ follow their gradient $\delta = -\gamma\D\cL$. We want to ask how much some parameters of interest $P \in \R^{d_P}$ contributed to that loss decrease. Decomposing the parameters as $\theta = (P; R)$, where $R \in \R^{d_R}$ contains the remaining parameters, the gradient partitions as $\D\cL = (\nabla_P \cL; \D_R \cL)$. We define the blockwise components of empirical Fisher
\[
    \F = \begin{pmatrix}
        \F_{PP} & \F_{PR} \\
        \F_{RP} & \F_{RR}
    \end{pmatrix}.
\]
In addition, for $A,B \in \{P,R\}$, we define
\[
    \alpha_A := \|\D_A\cL\|^2,
    \qquad
    C_{AB} := (\D_A\cL)^\top \F_{AB}\D_B\cL.
\]
Furthermore, we suppress the subscript $\theta$ in $D_\theta$ when clear from context.

\begin{proposition}[Exact loss contribution]
\label{prop:exact_param_cont}
    Let $\delta_{\gamma, u} = -(\gamma\D_P\cL;\, u)$ be a step direction. Assume $\F_{RR} \succ 0$.
    The contribution of $P$ to loss change~\eqref{eqn:quadratic} is
    \begin{align}
        D^{\textrm{exact}}_{P|R} &:= \min_u D(\delta_{0,u}) - \min_{\gamma, u} D(\delta_{\gamma,u}) \\
        &= \frac{1}{2}\frac{(\alpha_P - (\D_P \cL)^\top  \F_{PR} \F_{RR}^{-1} (\D_R \cL))^2}{C_{PP} - (\D_P \cL)^\top \F_{PR} \F_{RR}^{-1} \F_{RP} (\D_P \cL)},
    \end{align}
    provided the denominator is positive.
\end{proposition}
\begin{proof}
    For fixed $\gamma$, $D$ is minimised at $u^\star(\gamma) = \F_{RR}^{-1}(\D_R\cL - \gamma \F_{RP}\D_P\cL)$. Substituting it leaves $-\tfrac12(\D_R\cL)^\top \F_{RR}^{-1}\D_R\cL$, the best decrease of $R$ alone, plus a quadratic in $\gamma$, the minimisation of which gives the desired result.
\end{proof}

In practice, it is too computationally expensive to invert $\F_{RR}$, which is essentially the network-wide empirical Fisher. Instead, we compute a rank-1 version, which restricts $R$'s direction to its gradient while still allowing for an arbitrary magnitude.

\begin{corollary}[Gradient-restricted loss contribution]
\label{corr:restricted_param_cont}
    Restrict $u$ to the span of $\D_R\cL$, so that steps are $\delta = (-\gamma\D_P\cL; - \gamma_R\D_R\cL)$, $\gamma_R \in \mathbb R$. If $C_{RR} > 0$ and the denominator below is positive, the best such step decreases the quadratic model by at least as much as the best step with $\gamma = 0$, the difference being
    \vspace{-0.2cm}
    \begin{equation}
        D^\star_{P|R} = \frac{1}{2}\frac{\left(\alpha_P - \frac{C_{PR}}{C_{RR}} \alpha_R\right)^2}{C_{PP} - \frac{{C_{PR}}^2}{C_{RR}}}.
    \end{equation}
\end{corollary}
\begin{proof}
    \vspace{-0.2cm}
    As for Proposition~\ref{prop:exact_param_cont}, with $R$'s moves restricted to the span of $\D_R\cL$: $\D_R\cL$, $\F_{RP}\D_P\cL$ and $\F_{RR}$ are replaced by $\alpha_R$, $C_{PR}$ and $C_{RR}$.
\end{proof}

By restricting the updates of $P$ and $R$ to be along their gradients, we do not allow $R$'s update direction to change to compensate for $P$, however this gradient-restricted contribution needs only the five scalars $\alpha_P$, $\alpha_R$, $C_{PP}$, $C_{PR}$ and $C_{RR}$.

For a block $b$ with $m_b$ hidden channels, let $P_b$ denote the parameters of these channels and $R_b$ its complement. The credit $D^\star_{P_b \mid R_b}$ of block $b$, accumulated over the course of training (Section~\ref{sec:estimators}), measures only the contribution that its channels have made to the decrease in the loss. To turn this into a rule for \emph{where} to grow, we make two assumptions: i) after sufficient training, the recently-grown and previously-existing channels are interchangeable and ii) the loss contribution for $k$ channels is $k$ times that of a single channel.

At a growth event, let $k_b$ be the maximum number of neurons that can be added to block $b$ while respecting the budget increment $\Delta \mathcal{C}$. We grow the block $b$ with maximal value
\vspace{-0.2cm}
\begin{equation}\label{eqn:value_obj}
    b^\star = \operatorname*{argmax}_b \, k_b\, \frac{D^\star_{P_b|R_b}}{m_b}.
\end{equation}

\begin{remark}[Isotropic limit]\label{rem:gradmax}
  Neglecting the curvature $\F = I$, the objective~\eqref{eqn:value_obj} reduces to gradient maximisation $b^\star = \operatorname*{argmax}_b \frac{k_b}{m_b} \|\D_{P_b}\cL\|^2$, a budget-aware analogue of GradMax~\cite{evci_gradmax_2022}. Note however that GradMax only used its criterion to initialise the weights of new neurons, and not to determine where to grow.
\end{remark}

\section{Unbiased curvature estimates}\label{sec:estimators}
To estimate $D^\star_{P_b|R_b}$ for block $b$, we need $\alpha_{P_b}$, $\alpha_{R_b}$, $C_{P_bP_b}$, $C_{P_bR_b}$ and $C_{R_bR_b}$. However, the naive estimator is biased. As an example, let $g = \D_P\cL + \xi$ be an unbiased estimate of the gradient on $P$, with $\E [\xi] = 0$ and $\E [\xi \xi^\top] = \Sigma_P/N$, where $\Sigma_P$ is the covariance of one example's gradient on $P$ and $N$ the batch size. The estimate of $\|g\|^2$ over a minibatch is $\|\D_P \cL\|^2 + \operatorname{tr} \Sigma_P/N$, which is biased, in particular towards wide layers as $\operatorname{tr}\Sigma_P/N$ scales with the size of $P$.

Unbiased estimates of the components are obtained by estimating each factor on an independent minibatch. We use four consecutive minibatches drawn during training: the first two supply the gradient directions, and the last two the curvature. As the weights change slightly between these steps, the lemma below holds only approximately.

\begin{lemma} [Unbiased Fisher component estimate]
Let $g_1, \dots, g_4$ be the mean loss gradients on four independent minibatches of size $N$ drawn with replacement at the same weights, and let $g_{i,P}$ be the restriction of $g_i$ to $P$ (extended by zero). Let $\bar g = (g_3 + g_4)/2$, $\Delta = g_3 - g_4$ and $\lambda = (2N-1)/4$. Then $\widehat \F = \bar g\bar g^\top + \lambda\Delta\Delta^\top$ is an unbiased estimate of $\F$, and for any subsets of parameters $P$ and $Q$, $\hat\alpha_P = \inner{g_{1,P}}{g_{2,P}}, \widehat C_{PQ} = g_{1,P}^\top \widehat \F g_{2,Q}$ are unbiased estimates of $\alpha_P$ and $C_{PQ}$.
\end{lemma}
\begin{proof}
Recall that $\F = \E_x[q_x q_x^\top]$ where $q_x = \D \ell_x$. Write $\mu = \E[q_x] = \D\cL$ and $\Sigma = \operatorname{Cov}(q_x)$, so that $\F = \mu\mu^\top + \Sigma$. Since $\E[\bar g\bar g^\top] = \mu\mu^\top + \Sigma/(2N)$ and $\E[\Delta\Delta^\top] = 2\Sigma/N$, the choice of $\lambda$ above yields $\E[\widehat{\F}] = \F$. The remaining claims follow from independence of the batches.
\end{proof}
  
\begin{wrapfigure}{r}{0.43\textwidth}
  \vspace{-1.2cm}
  \begin{center}
    \includegraphics[width=\linewidth]{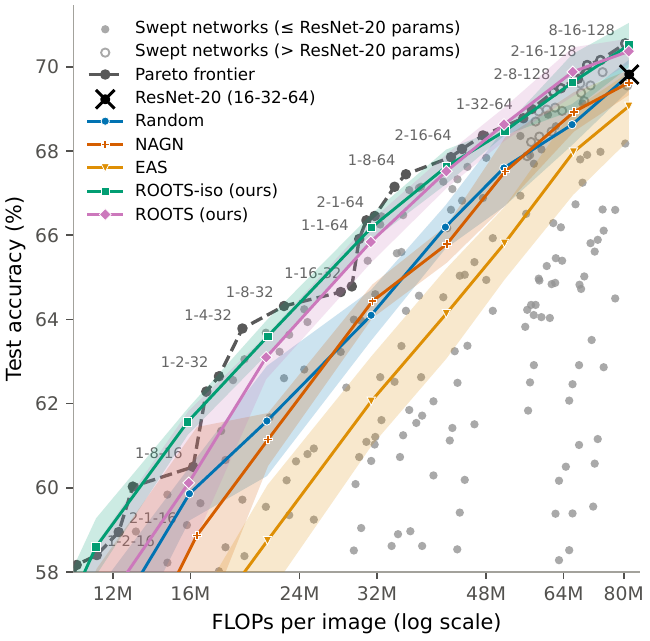}
    \caption{\small Sequence of grown architectures (retrained from scratch) vs the Pareto front of ResNet-20 channel distributions on CIFAR-100.}\label{fig:pareto}
  \end{center}
  \vspace{-2.9cm}
\end{wrapfigure}
Estimates are accumulated over training, with all quantities estimated over four consecutive training steps, then aggregated via an exponential moving average. The measurement adds no forward or backward passes. At a growth event, these quantities are then used to estimate the credit $D^\star_{P_b|R_b}$ associated with each block.

\section{Experiments}\label{sec:experiments}

\paragraph{Experimental Setup.}
The base architectures are VGG~\cite{simonyanZ14a} and ResNet-18/20 (CIFAR stem)~\cite{he_deep_2016}. The datasets considered are CIFAR-10/100 and Tiny\-ImageNet. Data augmentation and all other training hyperparameters follow~\cite{he_deep_2016}, except for the use of cosine annealing. Unless otherwise stated, all experiments use five seeds.

\paragraph{Growth mechanics}
For VGG we can grow the neurons of any of the convolutional layers, except the last one, which is connected to the linear head, starting at layer size 16. For ResNets (starting at layer size one), we do not grow the dimensions of the layers connected to the residual connection (similar to other growing methods~\cite{davey_survey_2026,evci_gradmax_2022}). Growth occurs once per epoch, for 100 epochs, with the complexity budget distributed evenly between events ($\Delta\C$ per event). The fully-grown network is then fine-tuned for a remaining 100 epochs. Each growth event is function preserving: random fan-in weights (Kaiming init) and zero fan-out. Growth, especially in the early stages, can dramatically change the size of layers, which can cause dead ReLUs. To alleviate this, we reset any neuron weights with identically-zero activations, at the beginning of each epoch, similarly to~\cite{sokar_dormant_2023}.

\begin{table}[t]
\centering
\caption{Final accuracy ($\to$ accuracy when retrained from scratch) reporting mean $\pm$ std over 5 seeds, parameter count and walltime (1 seed).}\label{tab:results}
\scriptsize\setlength{\tabcolsep}{1.5pt}
\makebox[\linewidth][c]{%
\begin{tabular}{@{}lr@{\;}c@{\;}lrrr@{\;}c@{\;}lrrr@{\;}c@{\;}lrr@{}}\toprule
 & \multicolumn{5}{c}{VGG-11 / CIFAR-10} & \multicolumn{5}{c}{ResNet-20 / CIFAR-100} & \multicolumn{5}{c}{ResNet-18 / Tiny-ImageNet}\\
\cmidrule(lr){2-6}\cmidrule(lr){7-11}\cmidrule(lr){12-16}
 Method & \multicolumn{3}{c}{Accuracy} & \llap{P (M)} & hrs & \multicolumn{3}{c}{Accuracy} & \llap{P (K)} & hrs & \multicolumn{3}{c}{Accuracy} & \llap{P (M)} & hrs\\
\midrule
 Baseline  & \multicolumn{3}{c}{92.6$\pm$.3} & 9.2 & .3 & \multicolumn{3}{c}{69.7$\pm$.2} & 280 & .4 & \multicolumn{3}{c}{66.2$\pm$.1} & 11 & 7.2 \\
 Random    & 92.4$\pm$.2  & $\to$ & 92.6$\pm$.2 & 9.2 & .3 & 69.7$\pm$.2 & $\to$ & 69.6$\pm$.5 & 300 & .3 & 66.4$\pm$.4 & $\to$ & 66.1$\pm$.4 & 12 & 6.1 \\
 NAGN      & 91.2$\pm$.2 & $\to$ & 91.3$\pm$.4 & 17. & .5 & 69.5$\pm$.5 & $\to$ & 69.6$\pm$.3 & 301 & .3 & 64.1$\pm$.2 & $\to$ & 63.1$\pm$.8 & 5.9 & 5.8 \\
 EAS       & 92.6$\pm$.2 & $\to$ & 92.6$\pm$.2 & 7.1 & .5 & 68.8$\pm$1. & $\to$ & 69.1$\pm$.8 & 250 & .7 & 62.6$\pm$.0 & $\to$ & 59.0$\pm$.1 & 2.7 & 5.3 \\ \midrule
 ROOTS-iso & 92.2$\pm$.4  & $\to$ & 92.5$\pm$.6 & 6.5 & .3 & 70.1$\pm$.4 & $\to$ & \textbf{70.5$\pm$.6} & 530 & .3 & \textbf{67.5$\pm$.5} & $\to$ & \textbf{67.5$\pm$.3} & 25 & 6.2\\
 ROOTS     & 92.3$\pm$.1  & $\to$ & \textbf{92.7$\pm$.2} & 6.2 & .4 & 70.4$\pm$.3 & $\to$ & 70.4$\pm$.3 & 380 & .4 & 67.4$\pm$.4 & $\to$ & 67.1$\pm$.1 & 18 & 5.7\\
\bottomrule
\end{tabular}
}
\end{table}

\begin{wrapfigure}{R}{0.45\textwidth}
  \vspace{-1cm}
  \begin{center}
    \includegraphics[width=\linewidth]{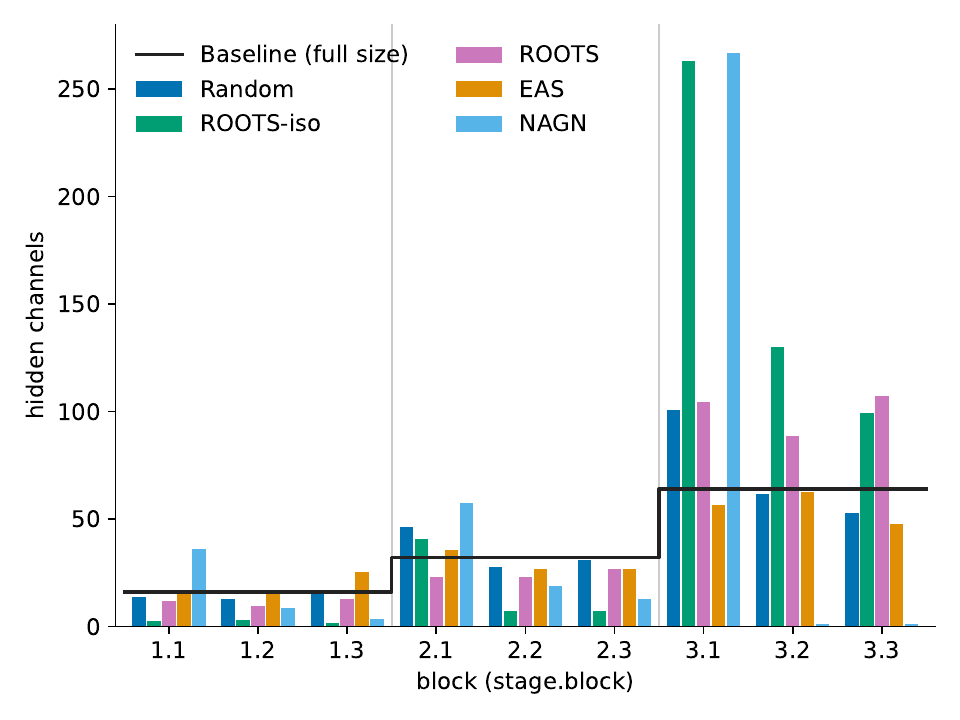}
      \caption{\small Final channel distributions on ResNet-20/CIFAR-100.}\label{fig:architecture}
  \end{center}
  \vspace{-0.8cm}
\end{wrapfigure}
\paragraph{Results.}
Table~\ref{tab:results} reports the accuracy of the grown networks, \emph{final acc} $\to$ \emph{retrained acc}, where in the latter case we reset the weights of the final architecture and train from scratch, to decouple the found architecture from the growing optimisation process. We compare to i) the fixed-architecture baseline, ii) budget-aware random block growth, iii) an isotropic $\F = I$ variant \methodiso (selecting $b^\star$ in Remark~\ref{rem:gradmax} at each growth step) of \method, iv) NAGN~\cite{wang_where_2026}, and v) EAS~\cite{wang_energy-aware_2020}.
To evaluate the sequence of architectures found during growth, we retrain a subset from scratch, shown in~\autoref{fig:pareto}. This is compared to an exhaustive grid search: for CIFAR-100, we train ResNet-20 from scratch for every stage-width triple in $\{1,2,4,8, \dots\}^3$, the three blocks of a stage sharing one hidden width, that fits within the standard network's FLOPs. Within the FLOP budget the most accurate is 8-16-128 ($70.55\pm0.1\%$) compared to the standard 16-32-64 ($69.82 \pm 0.4\%$).

Across all experiments, \method (and \methodiso) achieve the highest accuracy, matching or outperforming the baseline, with a considerable speedup for larger architectures/datasets. The isotropic variant \methodiso often achieves comparable performance to \method, however it requires significantly more parameters (\autoref{tab:results}), and learns highly unbalanced architectures (\autoref{fig:architecture}), compared to its curvature-aware method.

\section{Conclusion}\label{sec:results}

We designed a passive approach for estimating the contribution of a neuron during training: the credit of a block's existing channels, estimated from the gradients that ordinary training already computes, decides where new channels are added under a given FLOP budget. A single run of \method can yield a whole sequence of near-Pareto-optimal architectures at larger budgets, in comparable or less time than a standard training run. A small gap remains between the accuracy of the grown network and that of the same final architecture trained from scratch. We used a simple but strong baseline for new channels, random fan-in and zero fan-out weights; more sophisticated growth initialisations may be able to close this gap.

\begin{footnotesize}

\bibliographystyle{unsrt}
\bibliography{refs}

\end{footnotesize}

\end{document}